\documentclass[11pt,a4paper]{article}
\usepackage[hyperref]{acl2021}
\usepackage{times}
\usepackage{latexsym}

\usepackage{amsmath}
\usepackage{enumitem}
\usepackage{graphicx}
\usepackage{booktabs}
\usepackage{multirow}
\usepackage{tabularx}
\usepackage{microtype}
\usepackage{todonotes}
\usepackage{multirow}
\usepackage{subcaption}
\usepackage{amssymb}
\usepackage{amsmath}
\usepackage{array}
\usepackage{pgfplotstable}
\usepackage{pgfplots}
\usepackage{soul}
\usepackage{graphicx}
\usepackage{xcolor}

\usepackage{microtype}

\aclfinalcopy 

\newif\ifcomments
\commentstrue 
\ifcomments
    \providecommand{\dheeru}[1]{{\protect\color{magenta}{[Dheeru: #1]}}}
    \providecommand{\angela}[1]{{\protect\color{teal}{[Angela: #1]}}}
\else
    \providecommand{\dheeru}[1]{}
    \providecommand{\angela}[1]{}
\fi

\title{Tricks for Training Sparse Translation Models}

\def\authorspace{\hspace{4.5mm}}
\author{
Dheeru Dua\textsuperscript{$\heartsuit$}\authorspace{}
Shruti Bhosale\textsuperscript{$\spadesuit$}\authorspace{}
Vedanuj Goswami\textsuperscript{$\spadesuit$}\authorspace{}
James Cross\textsuperscript{$\spadesuit$}\authorspace{} \\
\textbf{Mike Lewis}\textsuperscript{$\spadesuit$}\authorspace{}
\textbf{Angela Fan}\textsuperscript{$\spadesuit$}\authorspace{}
\\
  \textsuperscript{$\heartsuit$}University of California, Irvine, USA \\
  \textsuperscript{$\spadesuit$}Facebook AI \\
  {\tt ddua@uci.edu} \\}

\date{}

\begin{document}
\maketitle
\begin{abstract}
Multi-task learning with an unbalanced data distribution skews model learning towards high resource tasks, especially when model capacity is fixed and fully shared across all tasks. Sparse scaling architectures, such as BASELayers, provide flexible mechanisms for different tasks to have a variable number of parameters, which can be useful to counterbalance skewed data distributions. We find that that sparse architectures for multilingual machine translation can perform poorly out of the box, and propose two straightforward techniques to mitigate this --- a temperature heating mechanism and dense pre-training. Overall, these methods improve performance on two multilingual translation benchmarks compared to standard BASELayers and Dense scaling baselines, and in combination, more than 2x model convergence speed.

\end{abstract}

\section{Introduction}

Training a universal model capable of handling many different tasks is a longstanding ambition in natural language processing~\cite{collobert2008unified,ruder2017overview,mccann2018natural}, with recent progress driven by training transformer models on a wide range of tasks~\cite{xue2020mt5,khashabi2020unifiedqa,lu202012}.
A central challenge in multi-task learning is accounting for the dramatically varying amounts of training data available for different tasks, which can lead to overfitting on low-resource tasks whilst simultaneously underfitting on tasks with abundant training data. 


In this work, we study multilingual machine translation as a multi-task learning problem~\cite{dong2015multi,domhan2017using}, where a single model is trained to translate between many language pairs \cite{fan2021beyond}. 
Multilingual learning has the potential of crosslingual transfer, allowing low-resource languages to benefit from high-resource data when trained together~\cite{conneau2019unsupervised}.
However, in practice, this positive transfer is often mitigated by interference between languages~\cite{arivazhagan2019massively,tan2019multilingual,zhang2020share}.
This is because all languages, irrespective of the amount of data, are trained with a fixed model capacity~\cite{lepikhin2020gshard}, leading to insufficient specialized capacity. 

Recent efforts have focused on sparse architectures~\cite{shazeer2017outrageously,lewis2021base} to train very high capacity models --- allowing high-resource languages sufficient specialization to reach stronger performance. 
However, these architectures can overfit to low-resource languages and often overall have worse performance than dense architectures~\cite{fan2021beyond,tran2021facebook}, which utilize all of their parameters for each training example.  
We analyze the learning patterns of experts throughout training and identify a fundamental problem: experts specialize early on and rarely change specialization. 

We propose two straightforward techniques to improve BASELayers-based sparse architectures~\cite{lewis2021base} for multitask learning: first, we slowly ramp the number of instances from low-resource tasks over epochs rather than having a fixed sampling ratio~\citep{arivazhagan2019massively}. This promotes cross-lingual transfer and reduces over-fitting as the model witnesses low-resource task instances in the later epochs.
Second, we train a dense architecture before switching to sparse training. Intuitively, we learn a generalized representation that can transfer across all tasks first with a dense model and then gradually sparsify and specialize the experts to different tasks.
Overall with these two modifications, we observe improvement in low-resource performance by 0.6 BLEU on WMT-15 benchmark and 1.1 BLEU on ML-50 benchmark --- whilst halving the training time.

\section{Methods}

We motivate the need for preventing early expert specialization and describe our proposed techniques to both circumvent this problem and more than double convergence speed.

\subsection{Expert Utilization Rarely Changes}

Sparse scaling schemes, such as BASELayers or Mixture-of-Experts, enable sparse computation by distributing model capacity across sets of experts. 
In each forward pass, only a small subset of experts are utilized, leading to incredibly compute-efficient scaling. 
The challenge, however, is the \textit{routing function} --- or how experts can be balanced so they actually specialize to learn different things~\cite{kudugunta2020exploring}. When the  routing mechanism is unbalanced, all the tasks degenerate to using only a single specific expert for all tasks~\cite{lepikhin2020gshard} --- essentially wasting parameters.
Often this problem is solved by penalizing the routing algorithm for un-utilized expert parameters via auxiliary losses. BASELayers~\cite{lewis2021base} employ a simple mechanism that learns a balanced routing without the need for additional auxiliary losses.
In our work, we focus on BASELayers as it has  straightforward and simple training and has previously been shown to have strong performance on language modeling tasks. 

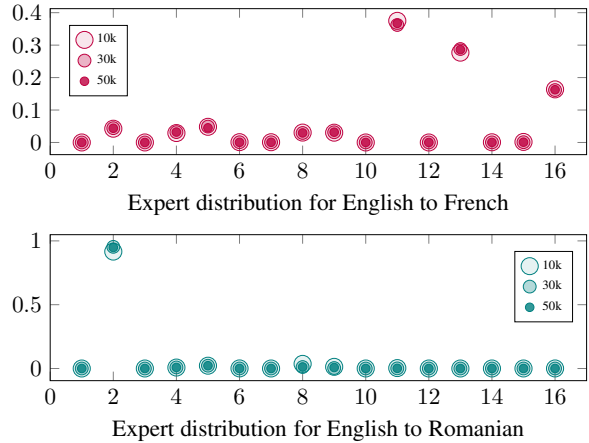
\begin{figure}[ht]
    \begin{subfigure}[t]{\textwidth}
    \begin{tikzpicture}[scale=0.8]
    \begin{axis}[
      width=0.65\textwidth, height=0.55*\axisdefaultheight,
      xlabel=Expert distribution for English to French,
      xmin=0,xmax=17,
      legend style={at={(0.09,0.4)},anchor=south,font=\tiny}]
    
    \addplot[only marks,color=purple,mark size=4,fill opacity=0.1] table [y=fr_10] {results/ro_fr_base.dat};
    \addlegendentry{10k}
    
    
    \addplot[only marks,color=purple,mark size=3,fill opacity=0.3] table [y=fr_30] {results/ro_fr_base.dat};
    \addlegendentry{30k}
    
    \addplot[only marks,color=purple,mark size=2,fill opacity=0.8] table [y=fr_50] {results/ro_fr_base.dat};
    \addlegendentry{50k}

    \end{axis}
    \end{tikzpicture}
    \end{subfigure}
    
    \begin{subfigure}[t]{\textwidth}
    \begin{tikzpicture}[scale=0.8]
    \begin{axis}[
      width=0.65\textwidth, height=0.55*\axisdefaultheight,
      xlabel=Expert distribution for English to Romanian,
      xmin=0,xmax=17,
      legend style={at={(0.92,0.4)},anchor=south,font=\tiny}]
    
    \addplot[only marks,color=teal,mark size=4,fill opacity=0.1] table [y=ro_10] {results/ro_fr_base.dat};
    \addlegendentry{10k}
   
    \addplot[only marks,color=teal,mark size=3,fill opacity=0.3] table [y=ro_30] {results/ro_fr_base.dat};
    \addlegendentry{30k}
    
    \addplot[only marks,color=teal,mark size=2,fill opacity=0.8] table [y=ro_50] {results/ro_fr_base.dat};
    \addlegendentry{50k}

    \end{axis}
    \end{tikzpicture}

    \end{subfigure}
    \caption{Expert distribution for Romanian and French as training progresses (10k, 30k and 50k updates) on WMT-15 benchmark, where Romanian is  low-resource and French is high-resource.}
    \label{fig:ro_fr}
\end{figure}

We observe that even though BASELayers leads to effective utilization of all parameters, it limits parameter sharing across tasks, which is crucial when the data distribution is unbalanced --- if the number of tasks and experts are the same, all tasks end up using a different set of experts. As a result, when applied to multilingual machine translation, the performance is worse than a corresponding dense architecture. Figure~\ref{fig:ro_fr} demonstrates that the main reason for limited parameter sharing is that expert assignment is fixed incredibly early on in training and rarely changes. 
Instead of learning how to better utilize capacity across high and low-resource languages over the training process, expert capacity is essentially frozen. 
We describe two strategies for more effective utilization of expert capacity, which can be easily applied to improve both low and high-resource translation performance.

\subsection{Balancing Low-Resource Tasks}

\paragraph{Temperature Sampling:}
To ensure that low-resource tasks are well represented during model training, temperature sampling~\citep{arivazhagan2019massively} is used to upsample low-resource tasks. If the data distribution across different tasks is $p$, then temperature sampling re-scales this distribution:
\begin{equation}
    p \leftarrow \frac{p_n^{1/T}}{\sum_{n \in |\text{tasks}|} p_n^{1/T}}
    \label{eq:temp_sampl}
\end{equation}

As we increase temperature from 1 to $\infty$, the sampling distribution changes from the original data distribution (e.g. highly skewed) to a uniform  distribution (e.g. tasks are equally represented). 

\paragraph{Temperature Heating:}
Instead of sampling data for each task with a fixed temperature at every epoch, we propose slowly increasing the temperature over the learning schedule  shown in Eq~\ref{eq:heating}. 
We define a starting temperature $t_s$, which is gradually increased at each epoch $e$, with a square root factor defined over maximum number of epochs $C$. The conduction coefficient $k$ determines the rate at which the temperature is increased.
\begin{equation}
    t_{e+1} = \sqrt{ \big( 1 + k\frac{e}{\sqrt{C}} \big) *  t_{s}^2 }
    \label{eq:heating}
\end{equation}

During the initial steps of training, this trains with lower temperatures, meaning high-resource tasks are better represented than low-resource tasks. 
As a result, the experts are more uniformly assigned across high-resource tasks. 
Upon slowly introducing low-resource tasks by increasing temperature during the learning process, the gating mechanism learns to route low-resource tasks through experts which were initially trained with high-resource tasks. 
This promotes positive cross-lingual transfer from high-resource languages to linguistically similar low-resource languages. 

\subsection{Dense Pre-training}
Architecturally, the sparsity in the output feed-forward layer of the transformer block can be viewed as a version of the same transformer on multiple GPUs with two main differences: the sparse feed-forward layers do not share parameters (have different initialization and gradients) and an additional gating mechanism decides which token should be routed to which expert. The alternative dense architecture would fully share parameters, so all parameters are utilized for each training example rather than routing to sparse parameters.

We propose first training a dense model for a fixed number of updates. Afterwards, we add a randomly initialized gating module and continue training the (output) feed-forward layers with sparsity, e.g. we do not average their gradients across compute nodes before back-propagating but update the weights individually in each node. 
As the sparse weights slowly diverge, they become more specialized towards specific tasks. 
Thus, models first learn a generalized representation when all parameters are fully shared, and then gradually specialize to handle different tasks.
Training in this fashion not only improves the learning of specialized experts, but also increases convergence.
\section{Experiments and Results}
\label{sec:results}

We demonstrate the advantages of our approach compared to BASELayers and dense baselines.
We experiment with $\mathtt{English} \rightarrow \mathtt{Many}$ multitasking on two benchmarks, WMT-15\footnote{http://www.statmt.org/wmt15/translation-task.html} and ML-50~\cite{tang2020multilingual} --- the first includes 15 languages and the second 50 languages. 
We use a Transformer~\cite{vaswani2017attention} sequence-to-sequence model with 6 encoder and decoder layers. We replace the final feed-forward layer of every alternate transformer block with a BASELayer. 
For ML50, we increase model capacity to 12 Transformer layers, following~\citet{tang2020multilingual}. 
We implement our methods in \texttt{fairseq}~\cite{ott2019fairseq} and evaluate performance with BLEU. 

\subsection{Effectiveness of Temperature Heating}

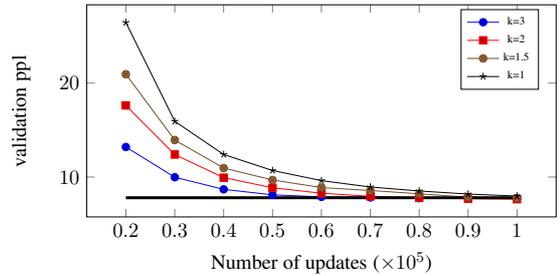
\begin{figure}[t]
    \centering
        \begin{tikzpicture}[scale=0.7]
        \begin{axis}[
          width=0.65\textwidth,
          height=0.35\textwidth,
          xlabel=Number of updates ($\times 10^5$),
          ylabel=validation ppl,
          legend style={at={(0.88,0.6)},anchor=south,font=\tiny}]
        \addplot table [y=3_all]{results/tmp_0_8_convg.dat};
        \addlegendentry{k=3}
        \addplot table [y=2_all]{results/tmp_0_8_convg.dat};
        \addlegendentry{k=2}
        \addplot table [y=1_5_all]{results/tmp_0_8_convg.dat};
        \addlegendentry{k=1.5}
        \addplot table [y=1_all]{results/tmp_0_8_convg.dat};
        \addlegendentry{k=1}
         \addplot[mark=none,black,ultra thick] coordinates {(0.2,7.8) (1,7.8)};
        \end{axis}
        \end{tikzpicture}
    \caption{Validation perplexity on ML-50 with $t_s$=0.8. Higher values of $k$ result in faster convergence.}
    \label{fig:time_convg}
\end{figure}

On WMT-15, training with BASELayers as a baseline has worse low-resource performance compared to a similarly sized dense model, losing 0.6 BLEU.
However, as we increase temperature, we recover the loss in low-resource task performance and also see improvements in the high-resource languages. The heating technique improves the overall BASELayers model performance by +0.7 BLEU (at $t_s$ = 0.8) (see Table~\ref{tab:temp_samp}). We observe similar trends in ML-50, where adding heating improves low-resource performance by +1.4 BLEU. Furthermore, temperature heating improves convergence speed. Given fixed $t_s$, the higher the $k$, the faster the model converges. As shown in Figure~\ref{fig:time_convg}, the model converges to same validation perplexity with $k$=3 at 50k updates as 100k updates with $k$=1.

\begin{table*}[htp]
\centering
    \small
    \begin{tabular}{l|lcccc}
    \toprule
    & \textbf{Model} & \textcolor{magenta}{Low-Resource} & & \textcolor{teal}{High-Resource} & All \\
     \midrule
    \bf WMT-15  & Dense & 13.3 & & 25.4 & 19.8  \\
        & BASELayers & 12.7 & & 25.3 & 19.4  \\
        & \hspace{0.5em} + heating ($t_s$=0.5) & 13.1 & & 26 & 20   \\
        & \hspace{0.5em} + heating ($t_s$=0.8) &  12.9 & & 26.4 & 20.1 \\
        &  \hspace{0.5em} + heating ($t_s$=1.0) &  13.2 & & 26.1 & 20.1 \\
        &  \hspace{0.5em} + heating ($t_s$=1.5) & 13.1 & & 26.1 & 20.0 \\
        &  \hspace{0.5em} + heating ($t_s$=2.0) & 13.3 & & 25.5 & 19.8  \\
    \midrule 
    & \textbf{Model} & \textcolor{magenta}{Low-Resource} & \textcolor{brown}{Mid-Resource} & \textcolor{teal}{High-Resource} & All \\
    \midrule
    \bf ML-50  & Dense  & 10.71 & 23.65 & 24.70 & 22.53 \\
     & BASELayers & 8.74 & 22.56 & 26.50 & 22.33 \\
     & \hspace{0.5em} +  heating ($t_s$=0.8) & 8.92 & 22.94 & 26.54 & 22.46 \\
     & \hspace{0.5em} +  heating ($t_s$=1.0) & 8.48 & 22.71 & 26.52 & 22.27 \\
     & \hspace{0.5em} +  heating ($t_s$=1.5) & 9.28 & 22.93 & 26.50 & 22.39 \\
     & \hspace{0.5em} +  heating ($t_s$=2.0) & 10.14 & 23.72 & 25.99 & 22.93 \\
    \bottomrule
    \end{tabular}
    \caption{Average BLEU and wall clock training time (until convergence) for different starting temperatures with a fixed conduction coefficient, $k$=1. The baselines are from our best performing dense and BASELayers models}
    \label{tab:temp_samp}
\end{table*}

\begin{figure}[htp]
\centering
   \begin{subfigure}[t]{\textwidth}
    \begin{tikzpicture}[scale=0.65]
    \begin{axis}[
      width=0.75\textwidth, height=0.3\textwidth,
      xlabel=Expert distribution with fixed temperature sampling,
      xmin=0,xmax=17,
      legend style={at={(0.1,0.6)},anchor=south,font=\tiny}]
    
    \addplot[only marks,color=violet,mark size=4,fill opacity=0.1] table [y=base_1] {results/expert_heating.dat};
    \addlegendentry{Epoch=1}
   
    \addplot[only marks,color=violet,mark size=3,fill opacity=0.3] table [y=base_2] {results/expert_heating.dat};
    \addlegendentry{Epoch=2}
    
    \addplot[only marks,color=violet,mark size=3,fill opacity=0.3] table [y=base_3] {results/expert_heating.dat};
    \addlegendentry{Epoch=3}
    
    \end{axis}
    \end{tikzpicture}
    \end{subfigure}
    \begin{subfigure}[t]{\textwidth}
    \begin{tikzpicture}[scale=0.65]
    \begin{axis}[
      width=0.75\textwidth, height=0.3\textwidth,
      xlabel=Expert distribution with heating,
      xmin=0,xmax=17,
      legend style={at={(0.1,0.6)},anchor=south,font=\tiny}]
    
    \addplot[only marks,color=violet,mark size=3,fill opacity=0.1] table [y=heat_1] {results/expert_heating.dat};
    \addlegendentry{Epoch=1}
    
    \addplot[only marks,color=violet,mark size=2.5,fill opacity=0.3] table [y=heat_2] {results/expert_heating.dat};
    \addlegendentry{Epoch=2}
    
    \addplot[only marks,color=violet,mark size=2,fill opacity=0.5] table [y=heat_3] {results/expert_heating.dat};
    \addlegendentry{Epoch=3}

    \end{axis}
    \end{tikzpicture}
    \end{subfigure}
 
    \caption{Expert distribution for a low-resource task English to Latvian as temperature is increased for WMT-15 with 16 experts.}
    \label{fig:exp_distrib_heat}
\end{figure}
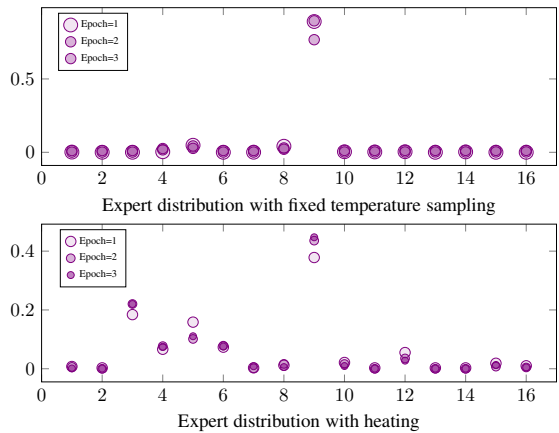

\subsection{Dense Pre-training with Heating}


For the WMT-15 benchmark, Table~\ref{tab:pretrain_results_wmt15} demonstrates that with dense pre-training, the best performing model improves by +0.75 BLEU over baseline BASELayers model but at the cost of 12 \% more computation time. 
To resolve this, we reduce computation time by introducing temperature heating, keeping the +0.7 BLEU improvement but reducing the computation time by $\sim$60\%. 

In the case of ML-50, results in Table~\ref{tab:pretrain_results_wmt15} confirm a similar trend. By combining Dense Pre-training with temperature heating, we improve +0.5 BLEU over a baseline BASELayers configuration and improve convergence speed by 2.5x. 
However, temperature heating can also be applied to the baselines. In those cases, on both benchmarks, we find that utilizing Dense Pre-training in combination with heating still has slightly better performance with significantly faster convergence. 

\subsection{Effect on Expert Distribution}
\begin{table*}[htp]
\centering
    \small
    \begin{tabular}{l|lcccc}
    \toprule
    & \textbf{Model} & \textcolor{teal}{High-Resource} & \textcolor{magenta}{Low-Resource} & All & Walltime (min)\\
     \midrule
    \bf WMT-15  & Dense & 25.2 & 12.7 & 19.4 & 76K  \\
        & \hspace{1em} + heating & 25.5 & 13.2 & 19.8 & 42K  \\
        & BASELayers & 25.2 & 12.9 & 19.5 & 77K \\
        & \hspace{1em} + heating & 26.0 & 13.0 & 19.9 & 42K  \\
        & Dense Pre-Train & 26.3 & 13.3 & \textbf{20.2} & 86K  \\
        & \hspace{1em} + heating & 26.0 & 13.4 & \textbf{20.1}  & \bf 31K \\
    \midrule 
    \bf ML-50  & Dense  & 24.7  & 10.7 & 22.5  & 596K \\
     & \hspace{1em} + heating & 25.9  & 10.5 & \bf 22.7 & 173K \\
     & BASELayers & 26.6   & 8.7 & 22.2  & 221K \\
     & \hspace{1em} + heating &  26.5  & 9.3 & 22.4 & 151K\\
     & Dense Pre-Train & 26.8  & 9.6 & 22.6 & 122K \\
     &  \hspace{1em} + heating  & 26.7  & 9.8 & \bf 22.7 & \bf 87K \\ 
    \bottomrule
    \end{tabular}
    \caption{Average BLEU on test set of WMT-15 and ML-50, with increasing number of dense pre-training steps at a starting/fixed temperature of 1.5. Wall clock time is the total training time including dense pre-training and sparse fine-tuning until the model reaches validation perplexity of 5.99 for WMT-15 and 7.6 for ML-50.}
    \label{tab:pretrain_results_wmt15}
\end{table*}

In standard BASELayer training, the learned expert distributions rarely change as the model trains (see Figure~\ref{fig:ro_fr}). 
This prevents expert capacity from being utilized more effectively, and contributes to low-resource overfitting.
In contrast, with our proposed techniques, the expert distribution changes and learns over training. 
Figure~\ref{fig:exp_distrib_heat} compares expert distribution between fixed temperature sampling and temperature heating over epochs for a low-resource language, demonstrating that temperature heating leads experts to change and learn over time.

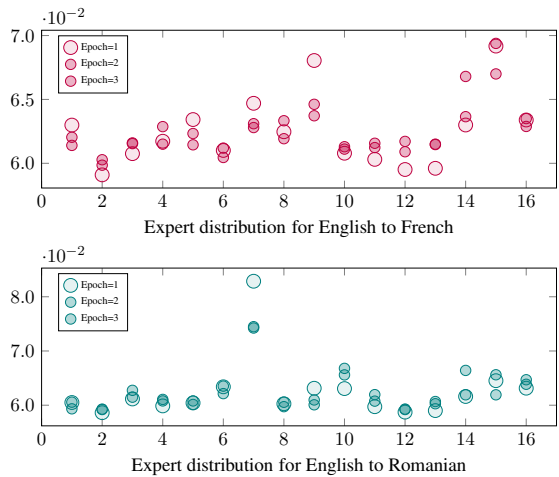
\begin{figure}[t]
    \begin{subfigure}[t]{\textwidth}
    \begin{tikzpicture}[scale=0.65]
    \begin{axis}[
      yticklabel=\pgfkeys{/pgf/number format/.cd,fixed,precision=1,zerofill}\pgfmathprintnumber{\tick},
      width=0.75\textwidth, 
      height=0.3\textwidth,
      xlabel=Expert distribution for English to French,
      xmin=0,xmax=17,
      legend style={at={(0.1,0.6)},anchor=south,font=\tiny}]
    
    \addplot[only marks,color=purple,mark size=4,fill opacity=0.1] table [y=fr_1] {results/expert_dense_pre_soft.dat};
    \addlegendentry{Epoch=1}
    
    \addplot[only marks,color=purple,mark size=3,fill opacity=0.3] table [y=fr_2] {results/expert_dense_pre_soft.dat};
    \addlegendentry{Epoch=2}
    
    \addplot[only marks,color=purple,mark size=3,fill opacity=0.3] table [y=fr_5] {results/expert_dense_pre_soft.dat};
    \addlegendentry{Epoch=3}

    \end{axis}
    \end{tikzpicture}
    \end{subfigure}
    
    \begin{subfigure}[t]{\textwidth}
    \begin{tikzpicture}[scale=0.65]
    \begin{axis}[
      yticklabel=\pgfkeys{/pgf/number format/.cd,fixed,precision=1,zerofill}\pgfmathprintnumber{\tick},
      width=0.75\textwidth, 
      height=0.3\textwidth,
      xlabel=Expert distribution for English to Romanian,
      xmin=0,xmax=17,
      legend style={at={(0.1,0.6)},anchor=south,font=\tiny}]
    
    \addplot[only marks,color=teal,mark size=4,fill opacity=0.1] table [y=ro_1] {results/expert_dense_pre_soft.dat};
    \addlegendentry{Epoch=1}
   
    \addplot[only marks,color=teal,mark size=3,fill opacity=0.3] table [y=ro_2] {results/expert_dense_pre_soft.dat};
    \addlegendentry{Epoch=2}
    
    \addplot[only marks,color=teal,mark size=3,fill opacity=0.3] table [y=ro_5] {results/expert_dense_pre_soft.dat};
    \addlegendentry{Epoch=3}

    \end{axis}
    \end{tikzpicture}

    \end{subfigure}
    \caption{Expert distribution for Romanian (low-resource) and French (high-resource) as sparse fine-tuning progresses on pre-trained dense model for WMT-15 with 16 experts.}
    \label{fig:exp_distrib_dense}
\end{figure}

In Figure~\ref{fig:exp_distrib_dense}, we show that by utilizing dense pre-training, we observe a high entropy in the expert distribution and increased expert sharing, indicating positive cross-lingual transfer from similar high to low-resource languages.

\section{Related Work}
\paragraph{Data Sampling} Low-resource tasks are  upsampled to balance their representation when pooled with high-resource tasks. Temperature sampling~\citep{arivazhagan2019massively} upsamples the data distribution based on a fixed temperature. This simple technique can result in over-fitting if the data distribution is skewed. Active learning sampling~\citep{gottumukkala2020dynamic} methods sample instances based on the current performance of task on dev set, which can be useful in case of catastrophic forgetting. Learned data samplers~\citep{wang2020balancing} can choose better sampling schemes but are computationally expensive.

\paragraph{Sparse Scaling} Sparsely-gated MoE models~\citep{shazeer2017outrageously} were introduced to increase model capacity in a flexible and scalable manner via model parallelism. The routing mechanism that decides which tasks should be routed to which set of experts is the key element that governs effective (better representation) and efficient (balanced assignment) resource utilization. To promote a balanced assignment, routing techniques~\citep{shazeer2017outrageously,lepikhin2020gshard,fedus2021switch} add a number of auxiliary task that encourage the routing mechanism to use a diverse set of experts. BASELayers~\citep{lewis2021base} circumvents this problem by treating the routing mechanism as a linear expert-to-task assignment problem, without the need of auxiliary loss. Routing networks~\citep{rosenbaum2017routing} learn better task representations by clustering and disentangling parameters conditioned on input.

\section{Conclusion}

We analyze the problem of balancing shared and specialized capacity in multitask learning, focusing on multilingual machine translation. We present two straightforward tricks to significantly increase convergence rate of mixture-of-expert models and improve their performance relative to dense baselines on two benchmarks.

\bibliographystyle{acl_natbib}
\bibliography{acl2021}
\clearpage
\appendix

\end{document}